\documentclass[sigconf,nonacm]{acmart}

\usepackage{amsfonts}
\usepackage{amsmath}
\usepackage{bbm}
\usepackage{booktabs} 
\usepackage[font=small]{caption}
\usepackage{enumitem}
\usepackage{graphicx}
\usepackage{makecell}  
\usepackage{multirow}
\usepackage{subfigure}
\usepackage[noend,ruled]{algorithm2e}
\usepackage{commath}  
\usepackage{tabularx}
\usepackage{dsfont}

\def\ddefloop#1{\ifx\ddefloop#1\else\ddef{#1}\expandafter\ddefloop\fi}
\def\ddef#1{\expandafter\def\csname bb#1\endcsname{\ensuremath{\mathbb{#1}}}}
\ddefloop ABCDEFGHIJKLMNOPQRSTUVWXYZ\ddefloop
\def\ddef#1{\expandafter\def\csname bf#1\endcsname{\ensuremath{\mathbf{#1}}}}
\ddefloop ABCDEFGHIJKLMNOPQRSTUVWXYZabcdefghijklmnopqrstuvwxyz\ddefloop
\def\ddef#1{\expandafter\def\csname c#1\endcsname{\ensuremath{\mathcal{#1}}}}
\ddefloop ABCDEFGHIJKLMNOPQRSTUVWXYZ\ddefloop
\def\ddef#1{\expandafter\def\csname v#1\endcsname{\ensuremath{\boldsymbol{#1}}}}
\ddefloop ABCDEFGHIJKLMNOPQRSTUVWXYZabcdefghijklmnopqrstuvwxyz\ddefloop
\def\ddef#1{\expandafter\def\csname v#1\endcsname{\ensuremath{\boldsymbol{\csname #1\endcsname}}}}
\ddefloop {alpha}{beta}{gamma}{delta}{epsilon}{varepsilon}{zeta}{eta}{theta}{vartheta}{iota}{kappa}{lambda}{mu}{nu}{xi}{pi}{varpi}{rho}{varrho}{sigma}{varsigma}{tau}{upsilon}{phi}{varphi}{chi}{psi}{omega}{Gamma}{Delta}{Theta}{Lambda}{Xi}{Pi}{Sigma}{varSigma}{Upsilon}{Phi}{Psi}{Omega}{ell}\ddefloop

\newcommand{\method}[1]{\mbox{\sf #1}\xspace}
\newcommand{\our}{\method{OA-Mine}}

\newtheorem{thm:def}{Definition}
\newtheorem{thm:eg}{Example}
\newtheorem{thm:lem}{Lemma}
\newtheorem{thm:obs}{Observation}
\newtheorem{thm:req}{Requirement}

\newcommand{\smallsection}[1]{\vspace{1mm}\noindent\textbf{#1.}}    

\setcopyright{acmcopyright}
\copyrightyear{2022}
\acmYear{2022}
\acmDOI{10.1145/1122445.1122456}

\copyrightyear{2022}
\acmYear{2022}
\setcopyright{acmcopyright}\acmConference[WWW '22]{Proceedings of the ACM Web
Conference 2022}{April 25--29, 2022}{Virtual Event, Lyon, France}
\acmBooktitle{Proceedings of the ACM Web Conference 2022 (WWW '22), April
25--29, 2022, Virtual Event, Lyon, France}
\acmPrice{15.00}
\acmDOI{10.1145/3485447.3512035}
\acmISBN{978-1-4503-9096-5/22/04}

\begin{document}


\title[\our: Open-World Attribute Mining for E-Commerce Products with Weak Supervision]{\our: Open-World Attribute Mining for E-Commerce Products with Weak Supervision}

\author{Xinyang Zhang$^{1}$, Chenwei Zhang$^{2}$, Xian Li$^{2}$, Xin Luna Dong$^{3}$\\ Jingbo Shang$^{4}$, Christos Faloutsos$^{5}$, Jiawei Han$^{1}$}
\affiliation{
    \institution{$^1$University of Illinois at Urbana-Champaign \quad $^2$Amazon.com, Inc. \quad $^3$Meta (Facebook) \\ $^4$University of California, San Diego \quad $^5$Carnegie Mellon University}
    \country{$^1$\{xz43,hanj\}@illinois.edu \quad $^2$\{cwzhang, xianlee\}@amazon.com \\ $^3$lunadong@fb.com \quad $^4$jshang@ucsd.edu \quad $^5$christos@cs.cmu.edu}
}

\renewcommand{\shortauthors}{X. Zhang et al.}

\begin{abstract}
\vspace{-3pt}
Automatic extraction of product attributes from their textual descriptions is essential for online shopper experience.
One inherent challenge of this task is the emerging nature of e-commerce products — we see new types of products with their unique set of new attributes constantly.
Most prior works on this matter mine new values for a set of known attributes but cannot handle new attributes that arose from constantly changing data.
In this work, we study the attribute mining problem in an open-world setting to extract novel attributes and their values.
Instead of providing comprehensive training data, the user only needs to provide a few examples for a few known attribute types as weak supervision.
We propose a principled framework that first generates attribute value candidates and then groups them into clusters of attributes.
The candidate generation step probes a pre-trained language model to extract phrases from product titles.
Then, an attribute-aware fine-tuning method optimizes a multitask objective and shapes the language model representation to be attribute-discriminative.
Finally, we discover new attributes and values through the self-ensemble of our framework, which handles the open-world challenge.
We run extensive experiments on a large distantly annotated development set and a gold standard human-annotated test set that we collected.
Our model significantly outperforms strong baselines and can generalize to unseen attributes and product types.

\vspace{-8pt}

\end{abstract}

\ccsdesc[500]{Information systems~Web mining}


\keywords{Open-world product attribute mining, weak supervision.}

\maketitle

\section{Introduction}
\label{sec:intro}

\begin{figure}[t]
    \centering
    \includegraphics[width=\linewidth]{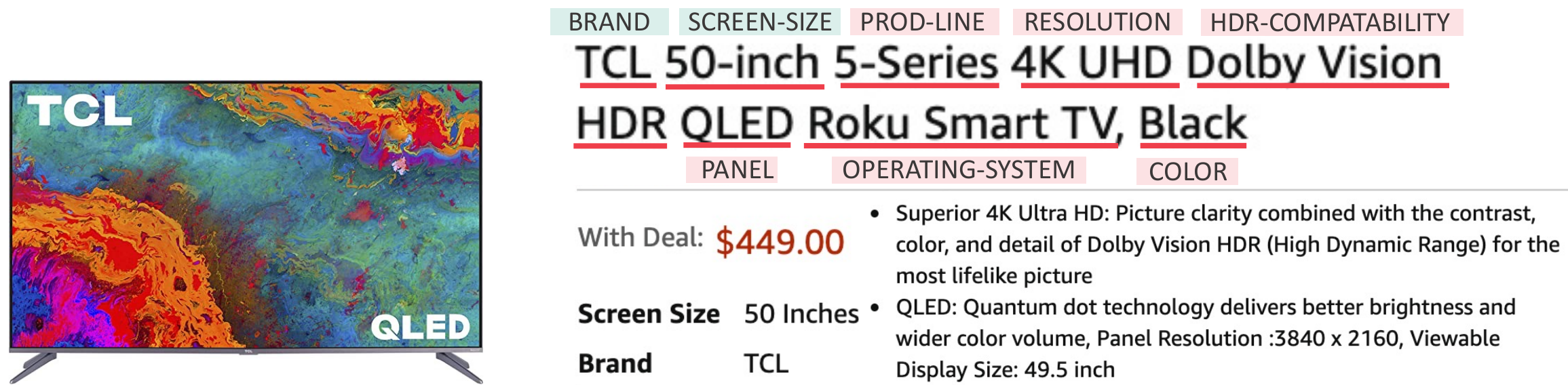}
    \caption{\small An example product title with known attributes annotated in green and new attributes in red. Sellers usually pack important product attribute values in the title for maximum exposure.}
    \label{fig:example_product}
\end{figure}



Product attributes are essential for online shoppers, as they provide valuable information to help search, recommendation, and product comparison.
The massive, constantly changing products bring the open-world challenge.
Products we already know may gain new attributes over time -- ``HDR compatibility'' barely comes to mind when people buy a TV ten years ago, but is relevant for many shoppers nowadays.
Not to mention, new types of products with distinctively new sets of attributes may emerge in the future.
Automatic extraction of product attributes has to meet the open-world challenge.
The model must be able to discover both \textit{attribute types} (e.g., brand, referred to as ``attributes'' hereafter) and \textit{attribute values} (e.g., Samsung, referred to as ``values'').
Moreover, they must operate with limited supervision, as human annotation can never keep up with the forever-expanding product catalog.

Existing works on attribute mining mostly carry a closed-world assumption on attributes.
Pioneer studies such as OpenTag~\cite{zheng2018opentag} bear ``open'' in the name, but their open-world assumption is on \textit{values only}.
They assume the set of attributes of interest is given as input to the model. 
The model is trained to find more values for the given attributes, but does not expand to unseen attributes without additional training data.
Moreover, they usually rely on extensive human labeling~\cite{zheng2018opentag} or noisy distantly supervised data~\cite{xu2019suopentag,yan2021adatag,wang2020aveqa}.

In this paper, we study the open-world attribute mining problem with weak supervision. 
We let the user provide a few example values for a few known attributes.
Neither the attributes nor the values given as examples are required to be comprehensive.
In addition, we also assume the type of the products are known a priori, as each product type holds a distinct set of applicable attributes (e.g., ``resolution'' applies to TVs but not shoes), and the attribute mining process should be done with respect to each product type.
With limited supervision and abundant product raw text, we aim to discover new attributes and values for products of different types.


Product titles, as the cleanest and most prominent part of product text, possess a number of unique properties.
As shown in Figure~\ref{fig:example_product}, product titles are concise, non-repeating, not written in a way as a general domain natural language.
More specifically, we highlight three observations from the data.
First, to maximize exposure of products to customers, sellers usually pack the highlights of their product in the title (observation 1: \textit{title first}).
Second, a product title rarely contains irrelevant information, and is a collection of attribute values (observation 2: \textit{bag-of-values}).
Third, with limited space in the title, the values seldom repeat (observation 3: \textit{value exclusiveness}).
For example, if one word describes the resolution of a TV product, the rest of the title will mention other attributes such as screen size.
These observations pave the way for a mining-based approach that exploits the concise product titles, is tailored to the language of products, and discovers new attributes and values in an open-world setting.

We propose \our, a principled framework for \textbf{O}pen-world product \textbf{A}ttribute \textbf{MIN(E)}ing with weak supervision.
Our framework has two main steps, \textit{candidate value generation} and \textit{attribute value grouping}.
The first step of our framework is designed based on the \textit{title first} and the \textit{bag-of-values} data observations.
It probes an in-domain fine-tuned language model that captures compositional relations between words, and segments product titles into phrases.
These phrases are candidates that will either be marked as an attribute value or be excluded as noise in our subsequent step.

\begin{figure}[t]
    \centering
    \subfigure[BERT trained w/ in-domain corpus]{
        \label{fig:first}%
        \includegraphics[width=0.44\linewidth]{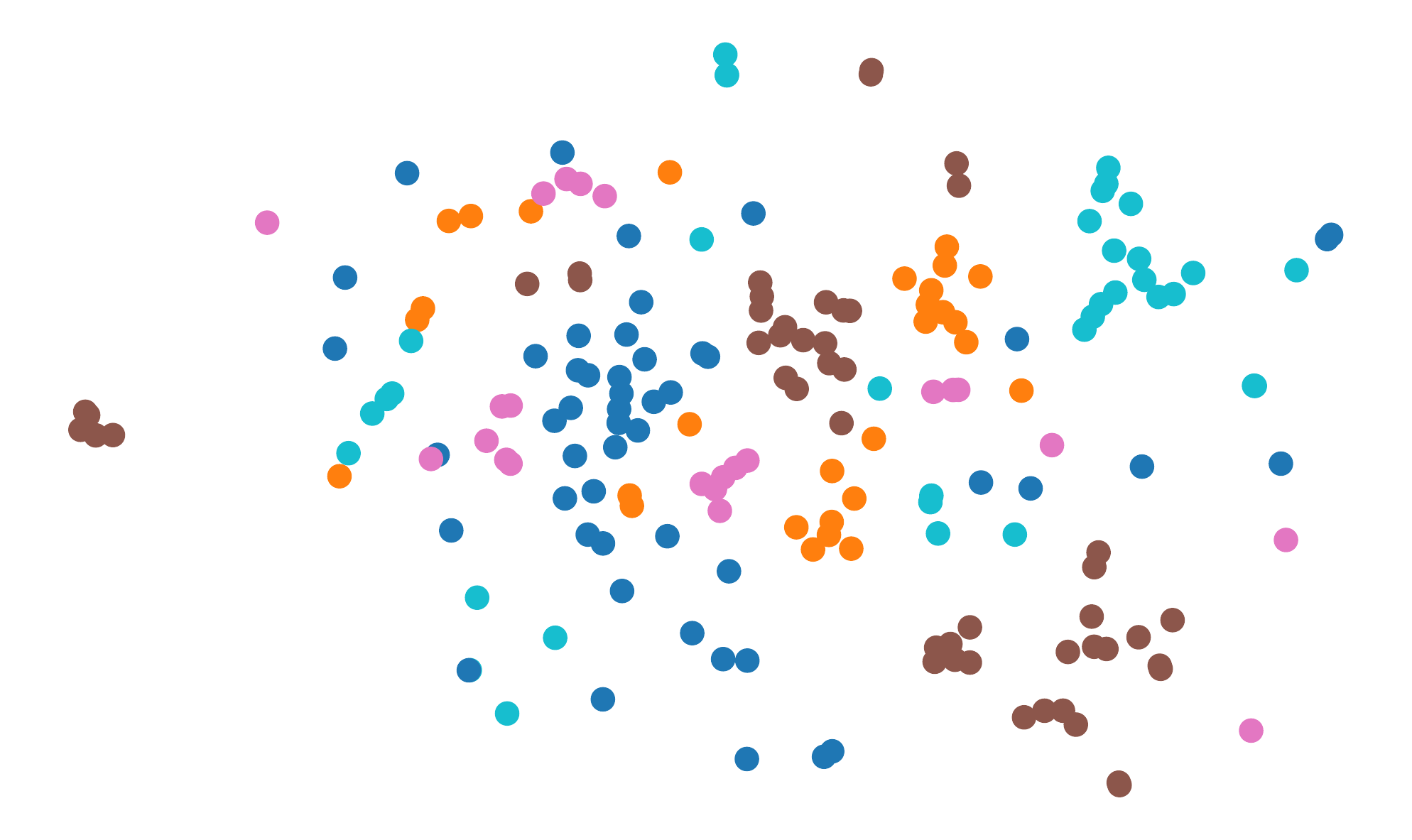}
    }
    \quad
    \subfigure[After attribute-aware fine-tuning]{
        \label{fig:second}
        \includegraphics[width=0.44\linewidth]{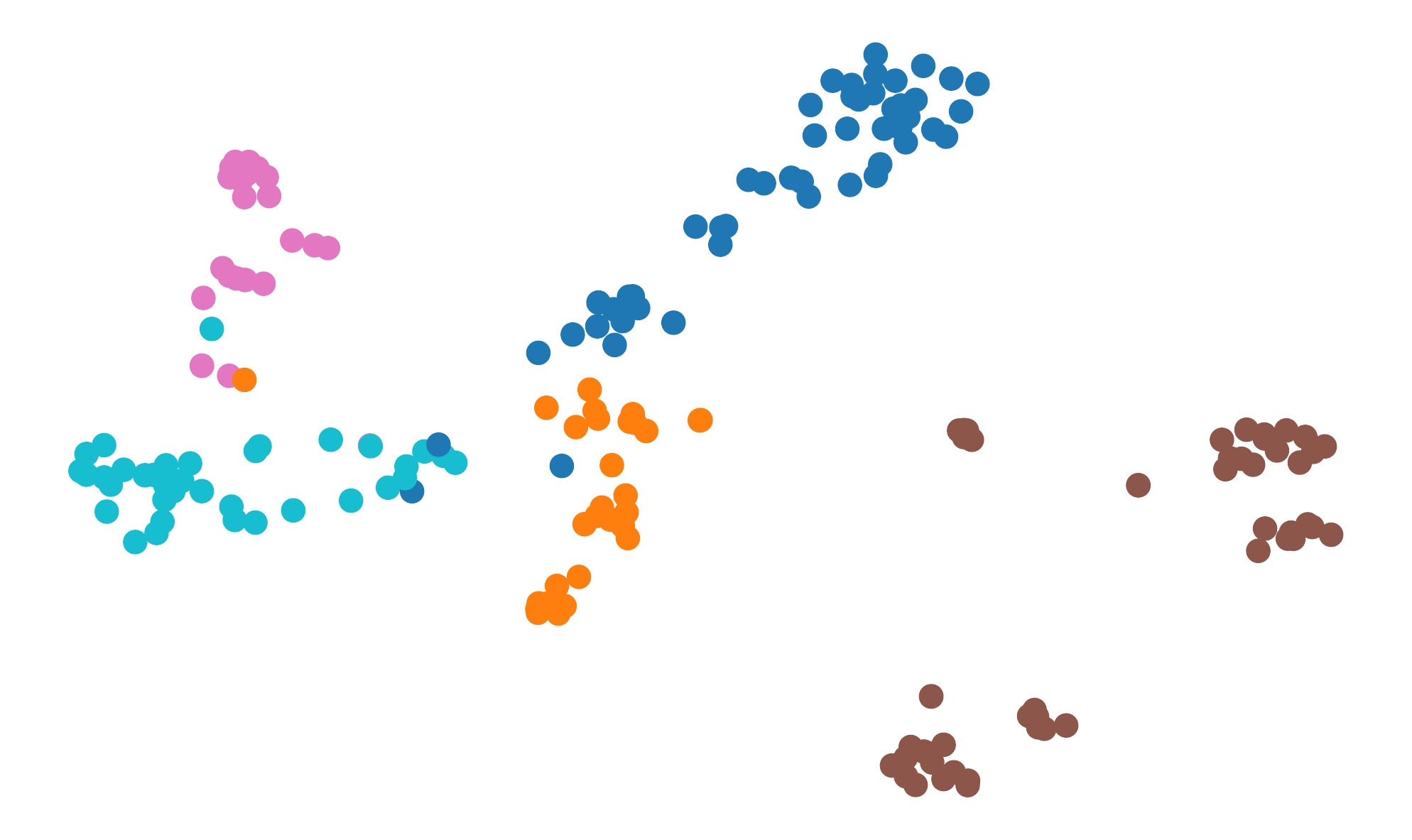}
    }
    \vspace{-5mm}
    \caption{T-SNE visualization of attribute value embedding. Points colored by attribute. After our attribute-aware fine-tuning, embedding is more reflective of attribute type.}
    \vspace{-3mm}
    \label{fig:embedding}
\end{figure}

Once we obtain the attribute value candidates, our framework finds the major attributes for each product type, and puts the candidates into different attribute groups.
We rely on a proper value embedding to propagate supervisions from seed values to other values when they together form a clique, and also to discover new values when new cliques emerge.
We find that pre-trained language model embedding (e.g., BERT embedding) cannot fully capture attribute-specific information, even after MLM~\cite{devlin2019bert} pre-training using in-domain corpus, as seen in Figure~\ref{fig:embedding}.
We propose an attribute-aware fine-tuning method, which takes full advantage of \textit{value exclusiveness} of product titles as well as the weak supervision, and optimizes a multitask objective function.
After the fine-tuning, embedding becomes more expressive in distinguishing values from different attributes.
With well-tuned embedding, we then apply a self-ensemble-based inference method.
A density-based clustering component, which can discover open-world new attributes and scrap noise from the candidates, is joint force with a classification component, which improves the model recall by finding more values of the discovered attributes.
Our \our framework improves itself with iterative training, where the discovered attribute types and values help gradually shape the value space for each attribute, as the model becomes more knowledgeable.


We run extensive experiments on over 750K products from 100 product types.
In addition to distantly annotated development data, we recruit crowd workers and domain experts to label a gold standard test set for end-to-end evaluation.
Our experiments show the superiority of our framework over various state-of-the-art baseline methods.
Besides, we also show the model's ability to discover unseen attributes by holding out attributes from training, and demonstrate the model's strong generalization by evaluating on product types with zero supervision.

Our main contributions can be summarized as follows:
\begin{itemize}[leftmargin=*,nosep]
    \item We propose a principled framework for open-world product attribute mining. The framework discovers both new attributes types and new attribute values.
	\item We propose a novel attribute-aware representation fine-tuning framework for pre-trained language models. The framework combines human given weak supervision with abundant unlabeled product text, and adapts the language model to distinguish values from different product attributes.
	\item We release a human annotated evaluation dataset for end-to-end evaluation of the open-world product attribute mining task.
\end{itemize}

\vspace{-2mm}
\section{Preliminaries}
In this section, we formally define the open-world product attribute mining task, and specify our input and output.
In addition, we offer some additional insights through data analysis. 
\begin{figure}[t]
    \centering
    \includegraphics[width=0.9\linewidth]{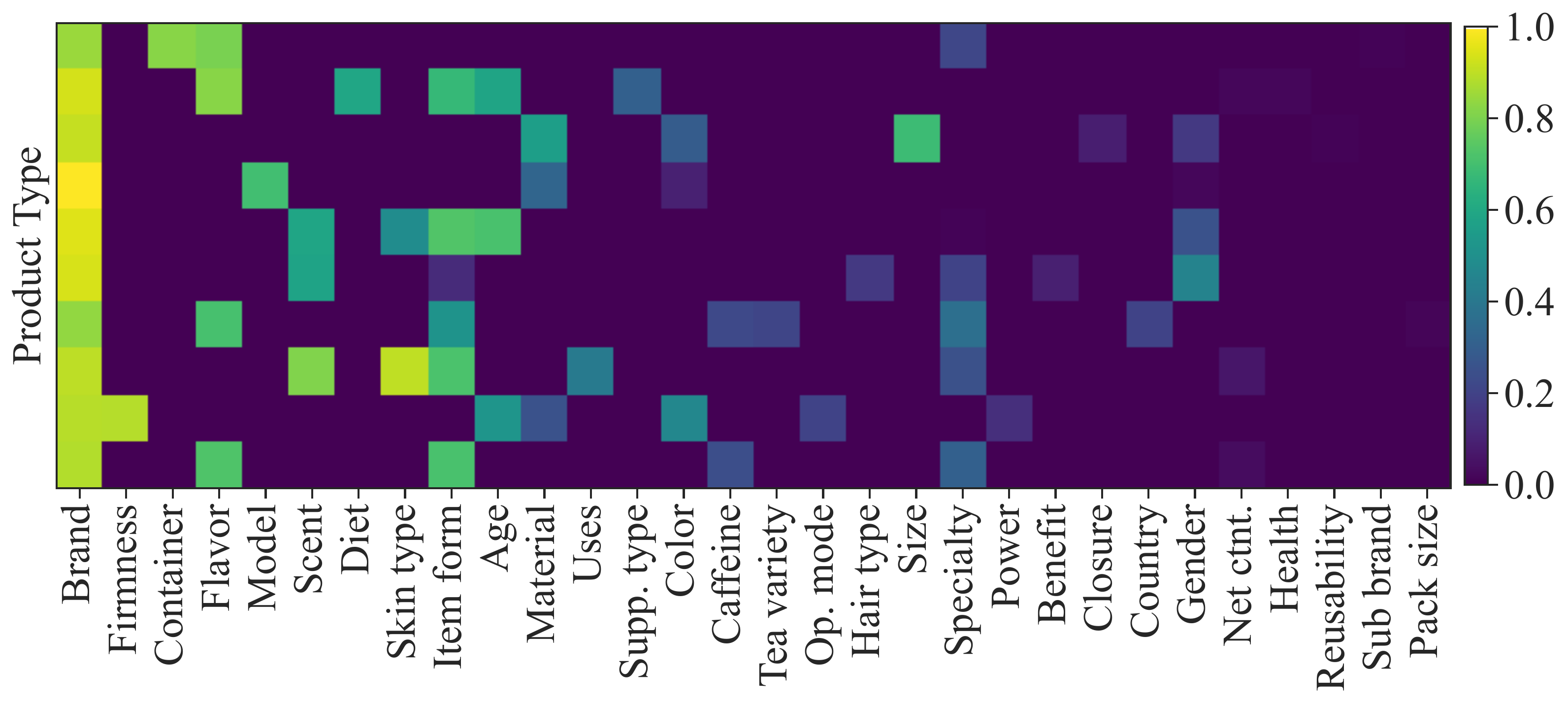}
    \vspace{-3mm}
    \small
    \caption{Attribute coverage. The cell color at $(x,y)$ is the coverage of attribute $x$ in product type $y$. $100\%$ means all values of this attribute are found in existing structured information in product profiles.}
    \vspace{-3mm}
    \label{fig:data_analysis}
\end{figure}

\smallsection{Problem Definition}
Open-world attribute mining for one type of product $t$ takes a set of products $\cP_t$ and aims to discover a set of applicable \textit{attributes} $\cA_t$ for those products.
Each applicable attribute $A \in \cA_t$ is represented by a collection of \textit{attribute values} $A =\{v_1, v_2, ... v_n\}$, which are raw text phrases.
The problem is guided by weak supervision, where the user specify a few known values $S_A=\{v_1, v_2, ..., v_k\}$ for a few known attributes $\cA_S \subset \cA_t$.
$S_A$ is called the seed set of attribute $A$.
The seed set size $|S_A|$ and the number of known attributes $|\cA_S|$ are small.
The predictions are attribute clusters $\bar{\cA_t}$ and the goal is to make the predictions $\bar{\cA_t}$ close to the ground truth $\cA_t$.

The problem is defined for a specific type of product $t$, because each product type has its unique sets of applicable attributes $\cA_t$.
Product types are identified by their surface names.
Putting all types of products together $\cP = \bigcup_{t\in T} \cP_t$, the open-world attribute mining 
aims to discover all the attributes $\cA_t$ for $t\in T$.

In this paper, we use product titles from $\cP$, and we assume the number of products $\envert{\cP}$ is large.

\begin{figure*}[t]
    \centering
    \includegraphics[width=\textwidth]{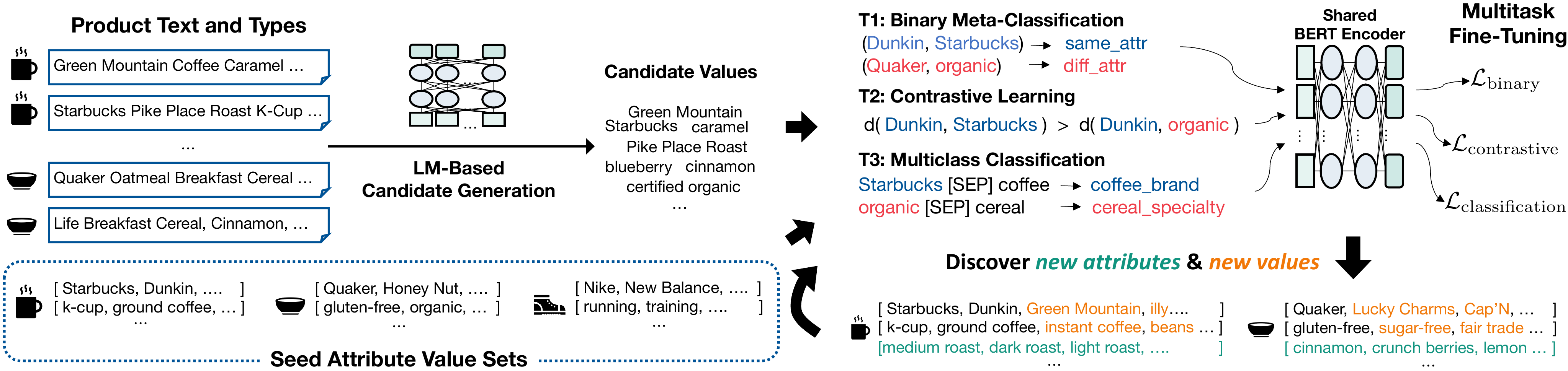}
    \vspace{-6mm}
    \caption{\small Overview of our proposed \our framework. Starting from product text, we generate candidate attribute values by probing a pre-trained language model (LM). Combined with user given seed sets, we fine-tune the LM with a multitask objective. Finally, we discover new attributes and values with self-ensembled based inference. Predictions are used for training in next framework iteration.}
  \vspace{-3mm}
    \label{fig:framework}
\end{figure*}

\smallsection{Data Analysis on Attribute Coverage}
To show the necessity of open-world product attribute mining, we collected 1,943 product profiles from 10 product types on Amazon.com for data analysis.
They contain raw text, as well as structured information, including some attribute values already known.
We hired human workers to annotate attribute values mentioned in product titles for a comprehensive analysis.
The details of the data collection process can be found in Section~\ref{sec:datasets}.

We would like to know what percentage of attribute values are already presented in the structured information of those online product profiles.
In total, the human annotators found 51 distinct attributes from all the products across 10 product types.
Of those, only 30 attributes are found in existing structured product profiles.
Figure~\ref{fig:data_analysis} shows the percentage of values identified by humans compared to the structured attribute information in product profiles.
Rows represent product types, and columns represent attributes.
Comparing different rows, we can see different types of products have different applicable attributes.
Judged by the color of the cells, lots of values are apparently missing.
The data analysis clearly shows the need for open-world attribute mining, as we see missing attributes and values in the existing product profiles.

\section{Framework Overview}

\our is a two-step framework for weakly supervised open-world product attribute mining (Figure~\ref{fig:framework}), with \textit{candidate value generation} followed by \textit{attribute value grouping}.
The first step aims to harvest phrases that are likely attribute values from raw text product titles.
Data observations from Section~\ref{sec:intro} show that product titles are collections of attribute values.
Our model segments the titles into phrases, essentially breaking down bags-of-attribute-values into individual value candidates.
Established phrase mining tools cannot meet our needs, as pre-trained NLP pipelines~\cite{spacy,akbik2019flair} are not adapted to the product text, while the unsupervised methods~\cite{gu2021ucphrase} typically has assumptions that do not work well on the product titles (shown in Section~\ref{sec:candidate_eval}).
We designed a language model probing-based technique, which captures the compositional relations between words for effective phrase segmentation.

Although product titles are primarily bags-of-attribute-values, there are exceptions where things that are not attribute values can be mentioned (e.g., ``packaging may vary'').
In our first step, we leave them as candidate values and let our second step remove them from the candidate pool.
We design our first step to be recall-focused and the second step to be noise-aware, as information lost from the first step cannot be recovered later, while noise can be effectively handled.

Having obtained the attribute value candidates, we find novel attributes and put the candidates into attribute groups in our next step.
Our model navigates the embedding feature space of candidate values and discovers dense areas that are likely attribute clusters.
Fundamentally, the embedding of candidate values has to be attribute-aware, a requirement we have shown that pre-trained language model embedding cannot satisfy.
To achieve this, we draw on the \textit{value exclusiveness} observation from data and combine it effectively with the limited supervision we have from seed values.
We optimize a multitask objective, with a binary meta-classification loss to generalize across product types, a contrastive loss to enforce embedding similarity resembles attribute-level similarity, and a classification objective to impose attribute distinctiveness.

Once we have well-trained embedding, a self-ensemble of different parts of the model first discovers new attributes and removes noise by DBSCAN~\cite{ester1996dbscan} clustering, and then puts more values into attribute groups with classification.
Finally, the model improves itself by leveraging the prediction for iterative training.



\vspace{-3mm}
\section{Candidate Value Generation}
\label{sec:candidate_generation}
    The first step of our framework aims to generate attribute value candidates from product titles.
    As product titles are mostly bags-of-attribute-values,  we formalize the task as a product title segmentation task.
    For example, for the product in Figure~\ref{fig:example_product}, we would like to segment its title into ``TCL | 50-inch | 5-Series | 4K UHD |...''
    Our goal of this step is to include as many value candidates as possible, i.e., it is recall-focused.
    We allow phrases that are not attribute values to be kept, but do not want actual attribute values to be lost.
    
    We leverage a pre-trained language model for candidate generation.
    As observed in prior work~\cite{kim2020lmphrase, gu2021ucphrase, wu2020perturbed}, pre-trained language models can capture phrases through the attention scores computed by the Transformer model.
    
    We first fine-tune a BERT~\cite{devlin2019bert} language model using the Masked Language Model (MLM)~\cite{devlin2019bert} objective on the product text.
    This step is necessary because the product text differs from the general domain natural language text on which BERT is pre-trained.
    Then we probe the BERT model to generate a score that resembles how likely two words are from the same phrase.
    
    \smallsection{Language Model Probing for Phrase Score}
        Given a product title as a sequence of words $W=w_1w_2...w_n$, we compute the likelihood $s(w_i,w_{i+1})$ of two adjacent words belonging to the same phrase.
        We use the following strategy~\cite{wu2020perturbed}:
        
        \begin{align}
            s(w_i, w_{i+1}) &= d\left( \mathrm{BERT}(W/\{w_i\})_i,  \mathrm{BERT}(W/\{w_i, w_{i+1}\})_i \right) \label{eqn:phrase_score}
        \end{align}
        
        $W/\{\cdot\}$ denotes the original text with words in the set under slash replaced with [MASK].
        $\mathrm{BERT}(\cdot)$ computes the sequence embedding by applying the BERT model, and $(\cdot)_i$ takes the i-th embedding, i.e., the token embedding of $w_i$.
        $d(\cdot, \cdot)$ is a distance function.
        We use cosine distance in our model.
    
    \smallsection{Phrasal Segmentation}
        With phrase score $s(w_i, w_{i+1})$ of each adjacent words computed, we can set a threshold for phrasal segmentation.
        Each pair of words with $s(w_i,w_{i+1})$ above the threshold will be merged into a phrase, and those below the threshold will be segmented into different phrases.
        We use a small validation set to select the threshold.
        Empirically we found the model and a single threshold generalize very well across different product types.
        
        One drawback of relying entirely on the language model is that it does not guarantee the completeness of the phrases.
        For example, ``QLED TV'' may be combined into one phrase in one product, but separated into two words in another product.
        This is because the BERT representation is context-dependent, and cannot leverage global frequency signals from the whole corpus.
        We, therefore, apply frequent pattern mining to combine words that frequently co-occur in the corpus, which further improves the completeness of the attribute value candidates.

\section{Attribute Value Grouping}
\label{sec:grouping}
    Once we have generated candidate values from product titles, we are ready to group them into attribute clusters.
    Given the open-world challenge in our problem setting, standard token classification approaches cannot suffice.
    Fully unsupervised clustering methods also fall short, as they rely on the pre-trained representation of phrases, which is not attribute-aware (Figure~\ref{fig:embedding}).
    
    We identify two main requirements for discovering new attributes and values:
    (1) The candidate value representation must be attribute-aware, placing values from the same attribute close to each other, and those from different attributes far apart;
    (2) The model must generalize to unseen attributes and to various product types.
    Bearing these two requirements in mind, we first design a multitask fine-tuning approach that focuses on attribute awareness and generalization.
    Then we apply a self-ensemble inference to discover new attributes and values.
    
    \subsection{Attribute-Aware Representation Fine-Tuning}
    \label{sec:fine-tune}
        Given the candidate attribute values, we use a shared encoder to obtain their embedding, and then we apply three different objective functions to fine-tune the model.
        
        \smallsection{Attribute Value Encoder}
        Given a candidate attribute value $v$ and its textual context in the product title $W$, we use a BERT~\cite{devlin2019bert} encoder to generate the token representation of the whole sequence.
        Next, we apply an average pooling on the token representations of the attribute value to obtain its representation $\vh_v$.
        Specifically,
        \begin{align}
            \hat{\vH} &= \sbr{\hat{\vh_1} \hat{\vh_2} ... \hat{\vh_n}} = \mathrm{BERT}(W) \\
            \vh_{v} &= \sum\nolimits_{w}  \del{ \hat{\vh_{w}} \cdot \mathds{1}\sbr{w \in v}} / \sum\nolimits_{w} \mathds{1}\sbr{w \in v}
        \end{align}
        The characteristic function $\mathds{1}\sbr{\cdot}$ equals to 1 for words in the candidate value, and 0 for the words in the context, thus masking out the hidden representation from the context.
        
        Recall that our input seed sets are collections of attribute values without context.
        When encoding the values from the seed sets, we string match them to their occurrences in the products, and use those occurrences for their contextualized representation.
        
        \smallsection{Binary Meta-Classification}
        The ultimate goal of fine-tuning is to embed values from the same attribute close to each other and embed values from different attributes far apart.
        We build a binary classification model $f$ to directly optimize for this goal: given a pair of values $v_1, v_2$, let $f(v_1, v_2)$ be $1$ if $v_1,v_2$ are from the same attribute, and $-1$ otherwise.
        
        We use a BERT bi-encoder~\cite{reimers2019sbert} with cosine similarity objective as the model $f$.
        \begin{align}
            f(u, v) &= \mathrm{cosine\_sim}\del{\vh_{u}, \vh_{v}} \\
            \cL_{\text{binary}} &= \sum_{(u,v)\in P} \enVert{1 - f(u,v)}^2 + \sum_{(u,v) \in N} \enVert{-1-f(u,v)}^2 \label{eq:l_binary}
        \end{align}
        $P$ and $N$ are positive and negative example sets respectively.
        We choose BERT bi-encoder as opposed to the standard BERT cross-encoder~\cite{devlin2019bert} because the bi-encoder is more efficient at inference time, as we will see in Section~\ref{sec:inference}.  
        
        Note that we would like the model to capture the attribute distinctiveness, not just on one attribute of one product type.
        Rather, we would like to build a \textit{meta-classifier} which generalizes to all attributes and all product types.
        We carefully sample the positive training pairs $P$ and the negative training pairs $N$ from both the seed sets and the unlabeled corpus.
        We generate positive pairs from values in the same attribute seed set, and negative pairs from different seed sets.
        In addition, based on the \textit{value exclusiveness} data observation, we generate strong negative training data by first sampling a product title from the corpus, and then sampling a pair of values from the same title.
        Our method generates higher quality pairs compared to random negative sampling.
        
        \smallsection{Contrastive Learning}
        In addition to directly optimizing the binary meta-classification loss, we find it beneficial to bind together the positive and the negative training pairs with a contrastive learning loss~\cite{schroff2015facenet,schultz2004distcompare,chechik2010imrank,chen2019simclr}.
        The contrastive learning objective takes a triplet of $(v_\mathrm{anc},v_\mathrm{pos},v_\mathrm{neg})$, and enforce the embedding distance from the anchor $v_\mathrm{anc}$ to the positive example $v_\mathrm{pos}$ to be smaller than that from the anchor $v_\mathrm{anc}$ to the negative example $v_\mathrm{neg}$.
        
        Contrastive learning has different instantiations in the choice of the loss function.
        Recent studies usually explicitly generate positive pairs with data augmentation but implicitly generate negative pairs~\cite{chen2019simclr}.
        Since we can explicitly generate strong negative pairs, we resort to a triplet loss function~\cite{schroff2015facenet}:
        \begin{align}
            \cL_{\mathrm{contrastive}} = \sum_{(v_a, v_p, v_n)} \max \del{\enVert{f(v_a, v_p)}^2 - \enVert{f(v_a, v_n)}^2 + \alpha, 0} \label{eq:l_contrastive}
        \end{align}
        $\alpha$ is a constant margin value. The triplets are generated by binding our positive pairs $P$ and negative pairs $N$.

        \smallsection{Multi-class Classification}
        The binary and the contrastive loss functions would shape the hidden feature space to be attribute-aware, as they manipulate the embedding space, so values that fall into the same attribute are pulled closer while values that belong to different types are pushed away from each other. 
        However, neither of those loss functions enforce the class distinctiveness of attribute values.
        Adding class distinctiveness by a multi-class classification loss would push the embedding from different attributes further apart.
        It offers additional benefits at inference time, shown in Section~\ref{sec:inference}, where it would improve the recall of the model.
        
        Let us assume we have already discovered the complete set of attributes.
        Now, we can build a multi-class classifier to put each value into an attribute.
        
        The challenge is scaling up to multiple product types, as each product type has a distinct set of applicable attributes.
        For example, the flavor of coffee can be very different from that of cereal.
        Most prior works do not distinguish product types and are suboptimal.

        We want to avoid building a separate classifier for each product type, while maintaining the product type awareness of the model.
        As such, we feed both the attribute value with context and the product type surface name to the multi-class classification model, delimited by the [SEP] special token.
        The output label space is the union of applicable attributes for all product types, e.g., ``coffee\_flavor'', ``tea\_brand''.
        We use the cross-entropy loss function for classification.
        \begin{align}
            \hat{\vy} &= \mathrm{Softmax}(\mathrm{Linear} (\mathrm{BERT}(W \text{[SEP]} t))) \\
            \cL_{\mathrm{classification}} &= \mathrm{CrossEntropy}(\hat{\vy}, \vy) \label{eq:l_clf}
        \end{align}
        
        $W$ is the attribute value $v$ in its context, $t$ is the product type surface name, and $\vy$ is the one-hot encoding of the class label.
        
        Note that we only use the attributes with seed values as our classes at the beginning of the training.
        We expand to more attributes through iterative training, where the predictions from the last iteration will reveal the complete attribute landscape.

    \subsection{Self-Ensemble Based Inference}
    \label{sec:inference}
        Once the model has been fine-tuned, we are ready to run an open-world inference to discover new attributes and values, given the attribute-aware embedding for each value.
        
        \smallsection{Attribute Discovery with DBSCAN}
        Thanks to the bi-encoder training with binary and contrastive objectives, the similarity of value embedding now reflects the likelihood of two values coming from the same attribute.
        
        We opt for DBSCAN~\cite{ester1996dbscan} to discover new attributes as it has several desirable properties.
        First, its local density requirement consolidates the attribute-level similarity computed by pairwise distance.
        Second, it discovers new attributes by finding dense cliques in the embedding space.
        Third, it excludes values that it deems noise if the embedding of a value is far from any dense clusters.
        
        
        \smallsection{Improving Recall with Classifier Inference}
        We run the multi-class classifier component after the fine-tuning.
        Each occurrence of a candidate attribute value is classified into an attribute cluster previously discovered by the DBSCAN model.
        The corpus level prediction of an attribute value is generated by majority voting of each occurrence of that value.
        
        \smallsection{Ensembling both Modules}
        Empirically, we have observed that the DBSCAN inference is very precision focused, leaving a large number of values out in a ``noise'' cluster.
        Whereas the classifier inference is recall-driven, expanding the model's coverage effectively.
        We, therefore, combine the attribute discovery ability of the DBSCAN inference with the classifier.
        Once DBSCAN discovered attribute clusters with high precision, we run the classifier on the noise cluster to include more values that were mistakenly excluded from the prediction.
        
        \subsection{Iterative Training}
        After one iteration of the framework, we have discovered new attributes and values.
        They are clusters of attribute values for different product types, and have the same format as our input seed sets.
        Therefore, we leverage the confident predictions as training data in the next iteration to boost the model performance further.
        
        There are two key benefits of iterative training.
        First, it augments the existing labeled seed sets, expanding the minimal supervision.
        Second, it offers a more complete landscape of attributes in the dataset.
        In our open-world setting, we do not have comprehensive supervision for all attributes in the beginning.
        The model predictions will alleviate this issue, especially for the classification objective in our framework, which operates on a set of known attributes.
        The complete training algorithm is shown in Algorithm~\ref{alg:tuning}.
        
\begin{algorithm}[t]
    \SetAlgoLined
    \LinesNumbered
    \small
    \KwIn{Seed sets $S$ for all product types $T$, candidate attribute values $C$, pre-trained model $\Theta$.}
    \Repeat{max\_iter}{
        \tcp{Attribute-Aware Representation Fine-Tuning}
        Initialize empty sets $D_{\text{binary}}, D_{\text{contrastive}},D_{\text{classification}}$ \;
        Initialize empty prediction set $P$ \;
        \For{$t \in T$}{
            $D_t \leftarrow $ $\text{generate\_data}(S_t \cup P_t, C_t)$ \tcp*[r]{Sec~\ref{sec:fine-tune}}
            Expand $D_{\text{binary}}, D_{\text{contrastive}},D_{\text{classification}}$ with $D_t$ \;
        }
        $D \leftarrow D_{\text{binary}} \cup D_{\text{contrastive}} \cup D_{\text{classification}}$ \;
        Shuffle and batch $D$ \;
        \For{$b \in D$}{
            Compute loss $L(\Theta)$ \\
            \quad $L(\Theta)=$ Eq.~\ref{eq:l_binary} for binary meta-classification \\
            \quad $L(\Theta)=$ Eq.~\ref{eq:l_contrastive} for contrastive learning \\
            \quad $L(\Theta)=$ Eq.~\ref{eq:l_clf} for classification \\
            Update model $\Theta \leftarrow \Theta - \epsilon \nabla(\Theta)$ \;
        }
        \tcp{Self-Ensemble Inference}
        $P \leftarrow \{\}$ \tcp*[l]{Empty previous predictions}
        \For{$t \in T$}{
            $P_t \leftarrow \text{run\_inference}(C_t, \Theta)$ \tcp*[r]{Sec~\ref{sec:inference}}
            $P \leftarrow P \cup P_t$  \;
        }
    }
    \caption{Attribute Value Grouping of \our}
    \label{alg:tuning}
\end{algorithm}


\section{Experiments}


\subsection{Datasets}
\label{sec:datasets}
\smallsection{Training Data}
We collect publicly available product profile data from Amazon.com for 100 different product types.  
Besides product titles, which will be used as textual descriptions of products, we also collect structured attributes found in these profiles (see Figure~\ref{fig:example_product} as an example).
For each product type, we find its applicable attributes, e.g., ``brand'', ``roast type'' for coffee and ``gender'', ``sports type'' for shoes.
For each applicable attribute, we examine the values mentioned in the product profiles that we collected, and take the most frequent values as the seed set for that attribute.
We keep up to 5 seed values for each attribute.

\smallsection{Test Label Set}
For end-to-end evaluation purposes, we manually selected products from 10 product types, sampled up to 200 products for each product type (1,943 products in total), and hired human workers for labeling.
We created a labeling tool that allows human workers to highlight attribute values mentioned in product titles and label their type. 
We defined a few applicable attributes we already know for each product type and added an option for human workers to label new attributes not seen in the predefined set.

First, each product is labeled by 5 crowd workers on Amazon Mechanical Turk.
Then, the labeled profiles are reviewed by a domain expert for consolidation.
After the values for known attributes are consolidated, we review the dataset for another pass and assign proper attribute types to those previously labeled as ``new attribute''.
On average, there are 11.5 attributes per product type marked by humans, and each attribute has 48.1 unique values.

The labels are gathered from human annotations, and are clusters of attribute values for the 10 selected product types. For example, labeled attribute values for coffee products are \{Brand: [Starbucks, Dunkin, ...], Roast Type: [medium roast, dark roast, ...], ...\}.

\smallsection{Development Label Set}
In addition to the human-annotated test set on selected 10 attribute types, we create a more extensive development set covering all 100 product types.
The labels are generated from the existing attribute information in the collected product profiles, using the same procedure described in the ``Training Data'' section for seed sets curation.
Aligned with training seed sets, we only keep values in the development when they are discovered by our candidate generation step.
On average, each attribute has 29.7 unique values.

With a limited budget, the development set would allow us to evaluate the performance of different models on 100 product types.
Note that for most prior works on product attribute mining~\cite{xu2019suopentag, wang2020aveqa, yan2021adatag}, the authors use the same method for gold-standard evaluation.
While in this paper, the development set serves the purpose of relative performance comparison.


\smallsection{Reproducibility}
Our code and data can be found at \url{https://www.github.com/xinyangz/OAMine}.\footnote{Note that although there were a few existing works~\cite{zheng2018opentag, xu2019suopentag, wang2020aveqa, yan2021adatag} on product attribute mining, only one of them~\cite{xu2019suopentag} provided a subset of their experiment data, and the dataset was labeled with distant supervision, thus do not serve our needs.}

\begin{table}[t]
    \center
    \caption{Evaluation on Attribute Value Candidate Generation. Methods are divided into pre-trained, distantly supervised, and unsupervised, from top to bottom.}
    \label{tbl:candidate_gen_results}
    \setlength\tabcolsep{2pt}
    \scalebox{0.95}{
        \begin{tabular}{l cccc}
            \toprule
            \textbf{Methods} & Entity-Prec. & Entity-Rec. & Entity-F1 & Corpus-Rec.\\
            \midrule
            spaCy~\cite{spacy} & 31.19 & 19.15 & 23.73 & 50.02\\
            FlairNLP~\cite{akbik2019flair} & 34.81 & 24.33 & 28.64 & 52.17 \\
            \midrule
            AutoPhrase~\cite{2018shangautophrase} & 26.58 & 29.67 & 28.04 & 32.39 \\
            \midrule
            UCPhrase~\cite{gu2021ucphrase} & 35.01 & 19.66 & 25.18 & 37.50 \\
            \our & \textbf{42.53} & \textbf{53.29} & \textbf{47.30} & \textbf{64.10} \\
            \bottomrule
        \end{tabular}
    }
\end{table}

\begin{table*}[t]
    \center
    \caption{End-to-end evaluation on development and test data. Results are average of 3 runs. Bold faced numbers indicate statistically significant results from t-test with 99\% confidence.}
    \vspace{-3mm}
    \label{tbl:e2e_results}
    \begin{tabular}{ll cccc cccc}
        \toprule
        & & \multicolumn{4}{c}{\textbf{Dev Set (100 product types)}} & \multicolumn{4}{c}{\textbf{Test Set (10 product types)}} \\
        \cmidrule(l){3-6} \cmidrule(l){7-10}
        \textbf{Method Type} & \textbf{Method} & ARI & Jaccard & NMI & Recall & ARI & Jaccard & NMI & Recall \\
        \midrule
        \multirowcell{3}[0pt][l]{Sequence tagging \\ (closed-world)} & BiLSTM-Tag & 0.299 & 0.354 & 0.422 &  0.565 & 0.175 & 0.219 & 0.374 & 0.162 \\
        & OpenTag~\cite{zheng2018opentag} & 0.244 & 0.324 & 0.334 & 0.593 & 0.160 & 0.247 & 0.357 & 0.165  \\
        & SU-OpenTag~\cite{xu2019suopentag} & 0.637 & 0.598 & 0.607 & 0.525 & 0.411 & 0.340 & 0.542 & 0.162 \\
        \midrule
        \multirowcell{2}[0pt][l]{Unsupervised clustering} & BERT+AG-Clus & 0.249 & 0.446 & 0.585 & 0.742 & 0.386 & 0.308 & 0.504 & \textbf{0.430} \\
        & BERT+DBSCAN & 0.133 & 0.146 & 0.507 & 0.131 & 0.385 & 0.412 & 0.575 & 0.186 \\
        \midrule
        \multirowcell{2}[0pt][l]{Weakly sup. clustering} & DeepAlign+~\cite{zhang2021deepaligned} & 0.175 & 0.226 & 0.336 & 0.729 & 0.257 & 0.208 & 0.426 & 0.389 \\
        & \our(no multitask) & 0.671 & 0.634 & 0.610 & 0.458 & 0.601 & 0.518 & 0.733 & 0.225  \\
        & \our & \textbf{0.704} & \textbf{0.689} & \textbf{0.629} & 0.747 & \textbf{0.712} & \textbf{0.650} & \textbf{0.781}  & 0.275 \\
        \bottomrule
    \end{tabular}
\end{table*}

\subsection{Candidate Generation Performance}
\label{sec:candidate_eval}
We compare the attribute value candidate generation module of our framework to the following baseline methods on the human-annotated test set:
\begin{itemize}[leftmargin=*,nosep]
        \item \textbf{spaCy~\cite{spacy}} is an industrial library with a pre-trained statistical noun phrase chunking model.
        \item \textbf{FlairNLP~\cite{akbik2019flair}} is a neural NLP library. We use its phrase chunking model~\footnote{\url{https://huggingface.co/flair/chunk-english}} that is based on a neural sequence tagger.
        \item \textbf{AutoPhrase~\cite{2018shangautophrase}} is a distantly-supervised phrase mining tool. It leverages a high quality phrase dictionary from Wikipedia and trains a statistical classifier based on POS-tags and statistical features of text spans.
        \item \textbf{UCPhrase~\cite{gu2021ucphrase}} is an unsupervised phrase mining framework. It first finds ``core phrases'' by mining frequent closed sequential patterns. It then uses the core phrases to train a CNN-based phrase classifier which leverage features from attention maps extracted from pre-trained language models.
\end{itemize}

\smallsection{Evaluation Metrics}
We report micro-averaged entity level precision, recall, F1, which are computed per instance.
In addition, we also report corpus-level recall where the predictions on all products are compared to the labeled set.

\smallsection{Results}
As shown in Table~\ref{tbl:candidate_gen_results}, our model consistently outperform the baseline methods by a large margin.
spaCy and FlairNLP are sub-optimal on our dataset because they are pre-trained on general domain corpora, with very different language characteristics than product titles.
AutoPhrase performs poorly as it lacks in domain distant training dictionary.
UCPhrase struggles on our dataset because (1) it can only capture multi-word phrases while attribute values can be single tokens, (2) the core phrases generated by pattern mining are low quality.
Unlike long documents on which the authors designed their model, product text is noisy and repetitive, undermining the pattern mining method.

\vspace{-2mm}
\subsection{End-to-End Evaluation}
The end-to-end evaluation runs each model on the complete training set, and compares their performance on the dev and test sets.

Note that there is no existing work that solves the open-world attribute mining problem to our knowledge.
We include several sequence tagging-based models, which are the de-facto models for closed-world product attribute mining.
We also combine the candidate generation results from our model with a few open-world classification and clustering models for evaluation.

\begin{itemize}[leftmargin=*,nosep]
        \item \textbf{BiLSTM-Tag} is a sequence tagging model with a bi-directional LSTM encoder. The model trains a stand alone tagging model for each known attribute in the seed set.
        \item \textbf{OpenTag~\cite{xu2019suopentag}} improves the BiLSTM-Tag model by adding an attention layer. The original model adopts a CRF layer for prediction. However, we found that the CRF layer consistently under-performs a linear layer in our weakly supervised setting. Therefore, we use a linear prediction layer instead.
        \item \textbf{SU-OpenTag\cite{xu2019suopentag}}\footnote{A more recent work~\cite{yan2021adatag} improves the model by innovating on the CRF layer. However, CRF does not perform well in our weakly supervised setting. Therefore, we exclude ~\cite{yan2021adatag} from comparison.} is an end-to-end product attribute mining framework. It encodes product text and attribute surface name separately with BERT~\cite{devlin2019bert}, and combines them with LSTM and attention layers. Unlike OpenTag, the model trains a single tagger for all the known attributes. We substitute the CRF layer in the original paper with a linear layer as it performs better.
        \item \textbf{BERT+AG-Clus} leverage BERT embedding for agglomerative hierarchical clustering. We use the attribute value candidates generated by our framework as the input to the clustering method.
        \item \textbf{BERT+DBSCAN} leverage BERT embedding for DBSCAN density-based clustering. We use the attribute value candidates generated by our framework as the input to the clustering method.
        \item \textbf{DeepAlign+~\cite{zhang2021deepaligned}} is an open-world intent classification framework with clustering and self training. We adapt it for open-world token classification.
        \item \textbf{OA-Mine (no multitask)} is an ablation of our model. It uses binary meta-classification as its only fine-tuning objective function. Meanwhile, it relies on DBSCAN only for inference, as the classifier component is turned off during training.
\end{itemize}

\smallsection{Experiment Setting}
For sequence tagging-based models that use distant supervision, following~\cite{xu2019suopentag}, we string match our weak supervision sets to the product text to obtain the annotated corpus.
The clustering models and the DeepAlign+ model can only handle open-world classification, but not value extraction.
As such, we use the candidate values generated by our model as the input to those models.
After model training, we run all the methods on the whole corpus for inference, and calculate the evaluation metrics on the product types from the development and test sets.
Therefore, our evaluation is transductive.
All models that utilize BERT use the same pre-trained model with in-domain MLM fine-tuning.

The main parameters of our model are candidate generation threshold, DBSCAN parameters, and number of iterations.
We set the candidate generation threshold as described in Section~\ref{sec:candidate_generation}.
The DBSCAN parameters are selected after our attribute-aware fine-tuning.
When generating fine-tuning data, a small validation set was held out.
The model's distance threshold for binary meta-classification performs well for DBSCAN, as they all rely on pairwise cosine distances.
We run iterative training until stable -- up to 5 iterations in practice.


\smallsection{Evaluation Metrics}
We compare the predicted attribute clusters with the ground truth clusters derived from human annotations.
We use standard clustering evaluation metrics, namely \textbf{adjusted Rand index (ARI)}, \textbf{Jaccard}, \textbf{normalized mutual information (NMI)} for clustering quality evaluation, and \textbf{recall} for clustering coverage evaluation. We refer readers to~\cite[Chapter~17]{zaki2020dmbook} for details of these metrics.
All metrics except for recall are \textit{computed on labeled attribute values} and micro-averaged across attributes and product types.
The range of ARI is $-1$ to $1$ and all the other metrics are $0$ to $1$, larger is better.

\smallsection{Results}
As shown in Table~\ref{tbl:e2e_results}, our model achieves the best performance across the board with the exception of recall on the test set.
Recall of BERT+AG-Clus and DeepAlign+ are superior because they cannot remove noise from the candidates, and include everything in their predictions.
We can clearly see the downsides of that, as the rest of the metrics indicating the quality of the predicted clusters are significantly lower than our model.


Sequence tagging-based models perform poorly for two main reasons.
First, they require more training data to work, and second, they cannot discover new attributes.
SU-OpenTag works better than the other two as it leverages BERT pre-trained embeddings.

Comparing our model to the BERT+DBSCAN model, we clearly see the necessity of attribute-aware fine-tuning.
In fact, without fine-tuning, DBSCAN is extremely sensitive to parameters and performs poorly despite our tremendous effort in tuning it to work.

The DeepAlign+ method is surprisingly worse than the AG-Clus unsupervised clustering in many metrics.
This is because the initialization that it based on is poor on our data, and the adaptive clustering component, i.e., self-training, fails to work on the poor initialization.
Moreover, the BERT representation used by DeepAlign+ is not attribute-aware.

Comparing our model to the ablation without multitask objective and self-ensemble inference, we can see performance increase across the board, with both better quality and better recall in our main model.
This demonstrates that multitask training boosts the attribute-awareness of the model, leading to better quality, while the self-ensemble generated better coverage.

\begin{table}[t]
    \center
    \caption{Performance on discovering new attributes. Experiment conducted with 5-fold cross-validation, where each fold holds out 20\% attributes from training.}
    \label{tbl:attr_cv}
    
    \scalebox{0.95}{
        \begin{tabular}{l cccc}
            \toprule
            \textbf{Methods} & ARI & Jaccard & NMI & Recall \\
            \midrule
            BERT+AG-Clus & 0.215 & 0.372 & 0.308 & 0.832 \\
            BERT+DBSCAN & 0.199  & 0.431 & 0.129 & 0.370 \\
            \midrule
            DeepAlign+ & 0.192 & 0.329 & 0.303 & 0.831 \\
            \our & 0.599 & 0.743 & 0.489 & 0.688 \\
            \bottomrule
        \end{tabular}
    }
\end{table}

\begin{table}[t]
    \center
    \vspace{-3mm}
    \caption{Comparing model predictions on unseen attributes during cross-validation. Red is error. }
    \vspace{-3mm}
    \label{tbl:case_study}
    \setlength\tabcolsep{2pt}
    \small
    \scalebox{0.95}{
        \begin{tabular}{l lp{2.2in}}
            \toprule
            \textbf{Attribute} & \textbf{Method} & \textbf{Predicted Cluster} \\
            \midrule
            &BERT+AG-Clus & green mountain, folgers, coffee fool, maxwell house, \textcolor{red}{coffee roasters}, nescafe, eight o clock, ... \\
            \cmidrule(l){2-3}
            \multirowcell{2}[0pt][l]{Coffee \\ Brand}&DeepAlign+ & \textcolor{red}{gourmet}, \textcolor{red}{keurig brewers}, starbucks, green mountain coffee, \textcolor{red}{donut}, dunkin donuts, ...   \\
            \cmidrule(l){2-3}
            &\our & starbucks, green mountain, folgers, coffee fool, maxwell house, nescafe, san marco coffee, ... \\
            \midrule
            \multirowcell{3}[0pt][l]{Laundry \\ Detergent \\ Form } &BERT+AG-Clus & powder, bottle, pacs, \textcolor{red}{original}, \textcolor{red}{2}, pods, \textcolor{red}{32 loads}, ... \\
            \cmidrule(l){2-3}
            & DeepAlign+ & liquid, \textcolor{red}{laundry}, \textcolor{red}{wash}, pack, \textcolor{red}{stain}, \textcolor{red}{natural}, ...   \\
            \cmidrule(l){2-3}
            &\our & liquid, powder, bottle, spray, carton, pods, soap,  ... \\
            \bottomrule
        \end{tabular}
    }
\end{table}

\vspace{-3mm}
\subsection{Evaluation on New Attributes}
To test model's ability on discovering new attributes, we run an experiment on the development set, as we have aligned attribute types for training seed sets and evaluation label sets.
For all the available seed sets in all product types, we hold out 20\% attribute types, and use the rest for training.
After training and inference, we test the model's performance on the 20\% held-out attributes.
We report the performance from 5-fold cross-validation on held out attributes.
We exclude the sequence labeling-based methods, as they are closed-world and cannot discover new attributes.

The results are shown in Table~\ref{tbl:attr_cv}.
We observe that our method generates significantly higher quality clusters measured by ARI, Jaccard, and NMI scores.
Note that BERT+AG-Clus and DeepAlign+ cannot exclude noise from the predictions, resulting in lower quality, albeit high recall.

Table~\ref{tbl:case_study} shows a case study of models' ability in discovering unseen attributes.
The selected attributes are held out during training.
The cluster that most resembles the target attribute is selected for each method.
Values are sorted by frequency.
We omit BERT+DBSCAN as it consistently under-performs our model.
As we can see, our model generates the most accurate and semantically coherent clusters.
BERT+AG-Clus and DeepAlign+ cannot remove noise from predictions, resulting in lower quality.
DeepAlign+ also suffers from semantic drift from adaptive clustering.

\vspace{-2mm}
\subsection{Evaluation on New Product Types}
We test the ability of our model to generalize to product types with no weak supervision seed sets.
In this experiment, we exclude the seed sets for product types in the test set, and use the remaining seed sets from the rest of the 90 product types for training.
The model has full access to the raw product text.
After training, we run inference on the products not seen during training, and compute the performance on the test set.
None of the existing attribute mining or open-world classification baselines can handle this challenging experiment setting.
We compare to the unsupervised clustering methods and also to our main experiment result.

The results are seen in Table~\ref{tbl:new_product_types}.
Compared to the unsupervised clustering, our model generates significantly higher quality predictions, as measured by ARI, Jaccard, and NMI.
Without access to supervision for these product types, our model performs understandably slightly worse than the full access model reported in the main experiment Table~\ref{tbl:e2e_results}.
The difference, however, is not large, showing the encouraging generalization of our model.

\begin{table}[t]
    \center
    \vspace{-3mm}
    \caption{Performance on new product types. Models tested on product types not seen during training.}
    \vspace{-3mm}
    \label{tbl:new_product_types}
    \scalebox{0.95}{
        \begin{tabular}{l cccc}
            \toprule
            \textbf{Methods} & ARI & Jaccard & NMI & Recall \\
            \midrule
            BERT+AG-Clus & 0.386 & 0.308 & 0.504 & 0.430  \\
            BERT+DBSCAN & 0.385 & 0.412 & 0.575 & 0.186 \\
            \midrule
            \our & 0.658 & 0.609 & 0.702 & 0.231  \\
            \bottomrule
        \end{tabular}
    }
    \vspace{-5mm}
\end{table}



\section{Related Work}

\smallsection{Product Attribute Mining}
Most prior work on product attribute mining formalize the problem as a sequence labeling task.
Inspired by named entity recognition models, earlier work leverage statistical models~\cite{putthividhya2011bootstrapped} for extraction.
With the advancement of deep learning, the most prominent systems designed in recent years adopt BiLSTM-CRF~\cite{zheng2018opentag, yan2021adatag} or BERT-BiLSTM-CRF~\cite{xu2019suopentag} architectures for attribute value extraction.
Supervised method combined with active learning~\cite{zheng2018opentag} was explored in OpenTag~\cite{zheng2018opentag}, while follow up works typically settle on distant supervision~\cite{xu2019suopentag, yan2021adatag, wang2020aveqa}.
Besides sequence labeling-based methods, AVEQA~\cite{wang2020aveqa} explored question answering-based models for attribute value extraction.

The major different between our work and the existing works is that we approach the problem with open-world assumptions on both attributes and values, while prior works assume the attributes of interest are given as input.
For supervision, we adopt weak supervision, which is a small amount of high quality data, instead of distant supervision, where a large amount of noisy data is used.

\smallsection{Open-World Text Classification}
Although few work studies entity extraction in the open-world setting, a few prior works have explored open-world text classification with different application scenarios.
Xu et al.~\cite{xu2019openprod} proposed a model for open-world product classification, where they learn a meta-classifier to either match a new example to a seen class or to assign it to a new class.
Zhang et al.~\cite{zhang2021deepaligned} studies open-intent discovery, which is essentially open-world text classification, with an adaptive clustering and self-training-based approach.
Most of these work relies on pre-trained language representation and adaptive clustering~\cite{xie2016dec}.

Our problem setting is more challenging than open-world text classification, as we have to extract attribute values from raw text.
In terms of methodology, our proposed model can take advantage of unique characteristics of the product text, and can learn better representation.
As seen in our experiments, the adaptive clustering-based methods do not work well out-of-the-box, as the product text is noisy and leads to poor convergence of adaptive clustering.

\section{Conclusion and Future Work}
In this paper, we proposed a principled framework for open-world product attribute mining.
The framework includes a novel candidate generation method, an attribute-aware representation fine-tuning model, and achieves open-world inference through self-ensemble.
Our framework offers strong performance and generalize to unseen attributes and product types.
In the future, we will extend our model to handle additional text fields besides product titles, such as using product introductions.
Improving the classifier to make open-world inference is also a promising direction.


\balance
\bibliographystyle{ACM-Reference-Format}
\bibliography{cited}

\end{document}